\documentclass[11pt,a4paper]{article}
\usepackage[hyperref]{acl2019}
\usepackage{times}
\usepackage{latexsym}

\usepackage{url}

\usepackage{booktabs}
\usepackage{amsmath}
\usepackage{todonotes}
\usepackage{comment}

\aclfinalcopy 
\def\aclpaperid{2216} 

\title{Self-Supervised Learning for Contextualized Extractive Summarization}

\author{
 Hong Wang$^\dagger$,
 Xin Wang$^\dagger$,
 Wenhan Xiong$^\dagger$,\\
 \textbf{Mo Yu$^{\ddag}$, 
 Xiaoxiao Guo$^{\ddag}$, 
 Shiyu Chang$^\ddag$,
 William Yang Wang$^\dagger$}
\\
 $^\dagger$ University of California, Santa Barbara\\
 $^\ddag$ IBM Research\\
 \{hongwang600, xwang, xwhan, william\}@cs.ucsb.edu,\\
 yum@us.ibm.com, \{xiaoxiao.guo, shiyu.chang\}@ibm.com  
 }

\date{}

\begin{document}
\maketitle
\begin{abstract}
  
   
Existing models for extractive summarization are usually trained from scratch with a cross-entropy loss, which does not explicitly capture the global context at the document level. 
    In this paper,
    we aim to improve this task by introducing three auxiliary pre-training tasks that learn to capture the document-level context in a self-supervised fashion. Experiments on the widely-used CNN/DM dataset validate the effectiveness of the proposed auxiliary tasks. Furthermore, we show that after pre-training, a clean model with simple building blocks is able to outperform previous state-of-the-art that are carefully designed. \footnote{Code can be found in this repository:
    \url{https://github.com/hongwang600/Summarization}}
    
\end{abstract}

\section{Introduction}


Extractive summarization aims at shortening the original article while retaining the key information through the way of selection sentences from the original articles. This paradigm has been proven effective by many previous systems \cite{DBLP:conf/sigir/CarbonellG98,DBLP:conf/emnlp/MihalceaT04,DBLP:conf/ecir/McDonald07,DBLP:conf/aaai/CaoWDLZ15}. In order to decide whether to choose a particular sentence, the system should have a global view of the document context, e.g., the subject and structure of the document.
However, previous works \cite{DBLP:conf/aaai/NallapatiZZ17,DBLP:journals/access/Al-SabahiZN18,DBLP:conf/acl/ZhaoZWYHZ18,DBLP:conf/emnlp/ZhangLWZ18} usually directly build an end-to-end training system to learn to choose sentences without explicitly modeling the document context, counting on that the system can automatically learn the document-level context.


\begin{figure}
\centering
\includegraphics[width=7cm]{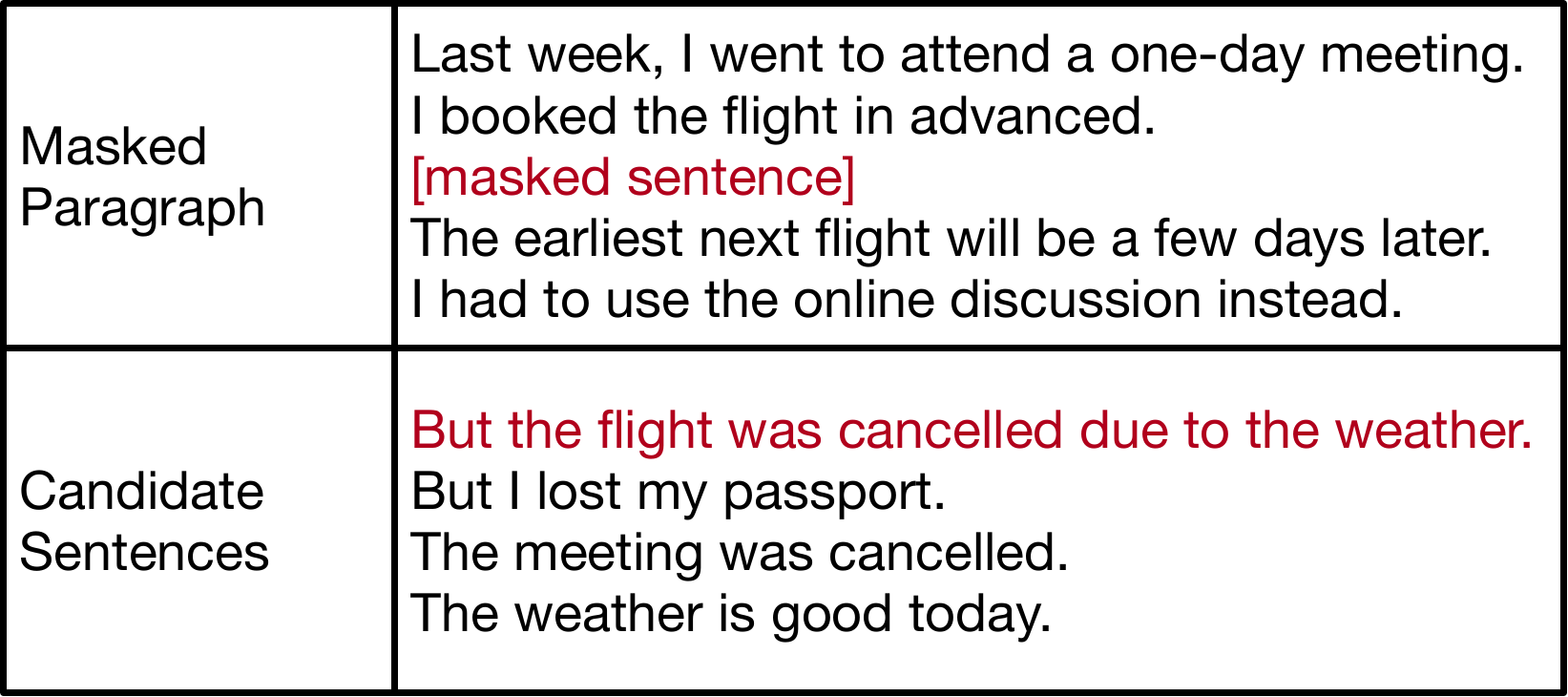}

\caption{
An example for the \textit{Mask} pre-training task. A sentence is masked in the original paragraph, and the model is required to predicted the missing sentence from the candidate sentences. 
}
\label{figure: example}
\vspace{-0.2in}
\end{figure}

We argue that it is hard for these end-to-end systems to learn to leverage the document context from scratch due to the challenges of this task, and a well pre-trained embedding model 
that incorporates 
document context should help on this task. In recent years, extensive works \cite{pennington2014glove,DBLP:conf/repeval/NieB17,DBLP:journals/corr/LinFSYXZB17,DBLP:conf/naacl/PetersNIGCLZ18,DBLP:journals/corr/abs-1810-04805,DBLP:journals/corr/abs-1804-00079,DBLP:journals/corr/abs-1803-11175,DBLP:journals/corr/abs-1803-02893,DBLP:conf/naacl/PagliardiniGJ18} have been done in learning the word or sentence representations, but most of them only use a sentence or a few sentences when learning the representation, and
the document context can hardly be included in the representation. Hence, we introduce new pre-training methods that take the whole document into consideration to learn the contextualized sentence representation with self-supervision. 


Self-supervised learning \cite{DBLP:conf/icml/RainaBLPN07,DBLP:conf/iccv/DoerschGE15,DBLP:conf/iccv/AgrawalCM15,DBLP:conf/iccv/WangG15} 
is a newly emerged paradigm, which aims to learn from the intrinsic structure of the raw data. The general framework is to construct training signals directly from the structured raw data, and use it to train the model. The structure information learned through the process can then be easily transformed and benefit other tasks. 
Thus self-supervised learning has been widely applied in structured data like text \cite{DBLP:conf/acl/OkanoharaT07,DBLP:conf/icml/CollobertW08,DBLP:conf/naacl/PetersNIGCLZ18,DBLP:journals/corr/abs-1810-04805, jiawie_ssl} and images \cite{DBLP:conf/iccv/DoerschGE15,DBLP:conf/iccv/AgrawalCM15,DBLP:conf/iccv/WangG15,DBLP:conf/iccv/LeeHS017}. 
Since documents are well organized and structured, it is intuitive to employ the power of self-supervised learning to learn the intrinsic structure of the document and model the document-level context for the summarization task.


In this paper, we propose three self-supervised tasks (\textit{Mask}, \textit{Replace} and \textit{Switch}), where the model is required to learn the document-level structure and context. The knowledge learned about the document during the pre-training process will be transferred and benefit on the summarization task.
Particularly, The \textit{Mask} task randomly masks some sentences and predicts the missing sentence from a candidate pool; The \textit{Replace} task randomly replaces some sentences with sentences from other documents and predicts if a sentence is replaced. The \textit{Switch} task switches some sentences within the same document and predicts if a sentence is switched. An illustrating example is shown in Figure \ref{figure: example}, where the model is required to take into account the document context in order to predict the missing sentence.
To verify the effectiveness of the proposed methods, we conduct experiments on the CNN/DM dataset \cite{DBLP:conf/nips/HermannKGEKSB15,DBLP:conf/conll/NallapatiZSGX16} based on a hierarchical model. We demonstrate that all of the three pre-training tasks perform better and converge faster than the basic model, one of which even outperforms the state-of-the-art extractive method \textsc{NeuSum} \cite{DBLP:conf/acl/ZhaoZWYHZ18}.

The contributions of this work include:

$\bullet$ To the best of our knowledge, we are the first to consider using the whole document to learn contextualized sentence representations with self-supervision and without any human annotations.

$\bullet$ We introduce and experiment with various self-supervised approaches for extractive summarization, one of which achieves the new state-of-the-art results with a basic hierarchical model.

$\bullet$ Benefiting from the self-supervised pre-training, the summarization model is more sample efficient and converges much faster than those trained from scratch.

\begin{figure}
\centering
\includegraphics[width=7cm]{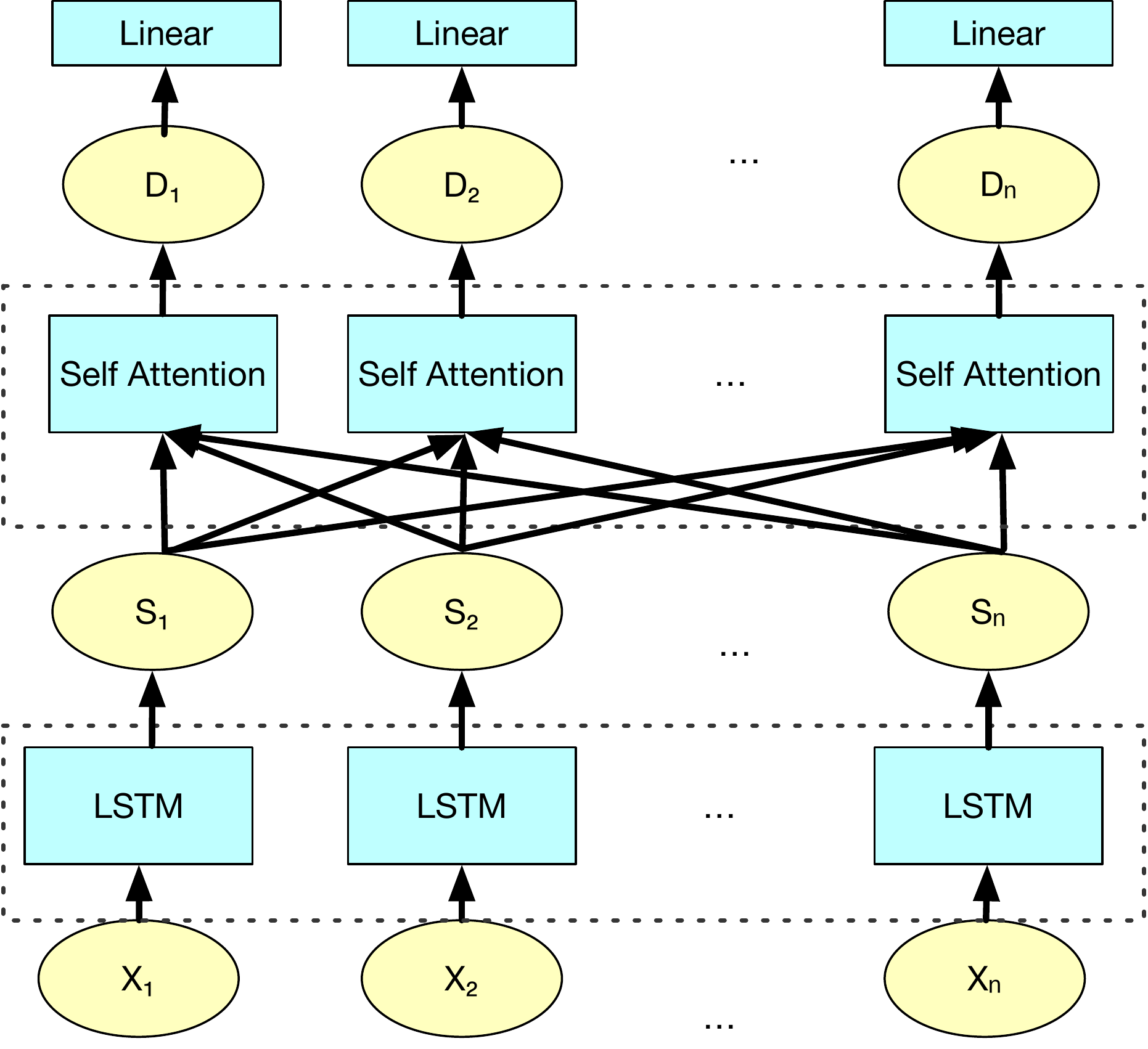}

\caption{The structure of the Basic Model. We use LSTM and self-attention module to encode the sentence and document respectively. $X_i$ represent the word embedding for sentence $i$. $S_i$ and $D_i$ represent the independent and document involved sentence embedding for sentence $i$ respectively.}
\label{figure: model}
\vspace{-0.1in}
\end{figure}

\section{Model and Pre-training Methods}
\subsection{Basic Model}
As shown in Figure \ref{figure: model}, our basic model for extractive summarization is mainly composed of two parts: a sentence encoder and a document-level self-attention module. 
The sentence encoder is a bidirectional LSTM~\cite{DBLP:journals/neco/HochreiterS97}, which encodes each individual sentence $X_i$ (a sequence of words) and whose output vector at the last step is viewed as the sentence representation $S_i$. 
Given the representations of all the sentences, a self-attention module~\cite{DBLP:conf/nips/VaswaniSPUJGKP17} is employed to incorporate document-level context and learn the contextualized sentence representation $D_i$ for each sentence.\footnote{We leave the combination of different architectures such as replacing the self-attention module with LSTM for future work.} Finally, a linear layer is applied to predict whether to choose the sentence to form the summary. 


\subsection{Self-supervised Pre-training Methods}


In this section, we will describe three self-supervised pre-training approaches.
Through solving each pre-training task, the model is expected to learn the document-level contextualized sentence embedding model from the raw documents, which will then be used to solve the downstream summarization task. Note that we are only pretraining the sentence encoder and document-level self-attention module of the basic model for extractive summarization. 

\paragraph{Mask}
Similar to the task of predicting missing word, the \textit{Mask} task is to predict the masked sentence from a candidate pool. Specifically, we first mask some sentences within a document with the probability $P_m$ and put these masked sentences ($\bf{x^m_1}, \bf{x^m_2}, \cdots, \bf{x^m_t}$) into a candidate pool $T^m$. 
The model is required to predict the correct sentence from the pool for each masked position $i$. 
We replace the sentence in the masked position $i$ with a special token \textit{$\langle$unk$\rangle$} and compute its document contextualized sentence embedding $D_i$. We use the same sentence encoder in the basic model to obtain the sentence embedding $S^m$ for these candidate sentences in $T^m$.
We score each candidate sentence $j$ in $T^m$ by using the cosine similarity:
$$
\Theta(i, j) = \cos (D_{i}, S^m_{j})
$$
To train the model, we adopt a ranking loss to maximize the margin between the gold sentence and other sentences:
$$
\ell_{m} = \text{max}\{0, \gamma - \Theta(i, j) + \Theta(i, k)\}
$$
where $\gamma$ is a tuned hyper-parameter, $j$ points to the gold sentence in $T^m$  for the masked position $i$, and $k$ points to another non-target sentence in $T^m$.

\paragraph{Replace}
The \textit{Replace} task is to randomly replace some sentences (with probability $P_r$) in the document with sentences from other documents, and then predict if a sentence is replaced. Particularly, we use sentences from $10,000$ randomly chosen documents to form a candidate pool $T^r$.
Each sentence in the document will be replaced with probability $P_r$ by a random sentence in $T^r$. Let $C_r$ be the set of positions where sentences are replaced. We use a linear layer $f_r$ to predict if the sentence is replaced based on the document embedding $D$, and minimize the MSE loss:
$$
\ell_r = \text{MSE}(f_r(D_i),~y^r_i)
$$
where $y^r_i=1$ if $i\in C_r$ (i.e., the sentence in position $i$ has been replaced), otherwise $y^r_i=0$.

\paragraph{Switch}
The \textit{Switch} task is similar to the \textit{Replace} task. Instead of filling these selected sentences with sentences out of the document, this task chooses to use sentences within the same document by switching these selected sentences, i.e., each selected sentence will be put in another position within the same document. Let $C_s$ be the set of positions where the sentences are switched. Similarly, we use a linear layer $f_s$ to predict if a sentence is switched and minimize the MSE loss: 
$$
\ell_s = \text{MSE}(f_s(D_i),~y^s_i)
$$
where $y^s_i=1$ if $i\in C_s$, otherwise $y^s_i=0$.


\section{Experiment}

To show the effectiveness of the pre-training method (\textbf{Mask}, \textbf{Replace} and \textbf{Switch}), we conduct experiments on the commonly used dataset CNN/DM \cite{DBLP:conf/nips/HermannKGEKSB15,DBLP:conf/conll/NallapatiZSGX16}, and compare them with a popular baseline \textbf{Lead3} \cite{DBLP:conf/acl/SeeLM17}, which selects first three sentences as the summary, and the state-of-the-art extractive summarization method \textbf{\textsc{NeuSum}} \cite{DBLP:conf/acl/ZhaoZWYHZ18}, which jointly scores and selects sentences using pointer network. 

\begin{figure}
\centering
\includegraphics[width=7.4cm]{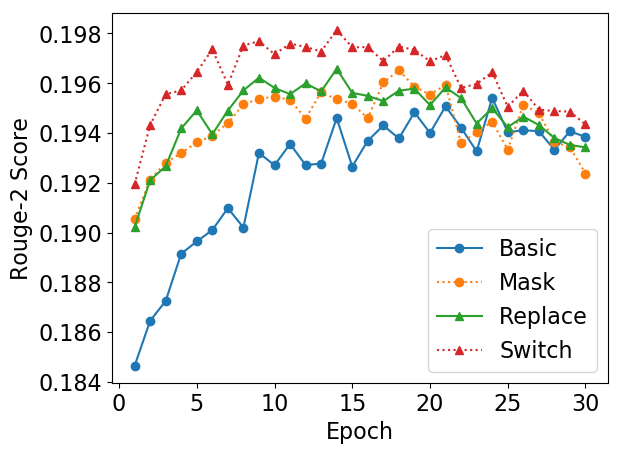}

\caption{This figure shows the Rouge-2 score for each pre-training method and the basic model on the development set during the training process. We put the result for Rouge-1 and Rouge-L score in Appendix \ref{rouge-1-L}}
\label{rouge score fig}
\vspace{-0.1in}
\end{figure}


\subsection{On CNN/DM Dataset}
\paragraph{Model and training details} We use the rule-based system from \cite{DBLP:conf/acl/ZhaoZWYHZ18} to label sentences in a document, e.g., sentences to be extracted will be labeled as $1$. Rouge score\footnote{We use PyRouge \url{https://pypi.org/project/pyrouge/} to compute the Rouge score.} \cite{lin2004rouge} is used to evaluate the performance of the model, and we report Rouge-1, Rouge-2, and Rouge-L as in prior work.
We use the pre-trained glove embedding \cite{pennington2014glove} with $100$ dimensions to initialize the word embedding. A one-layer bidirectional LSTM \cite{DBLP:journals/neco/HochreiterS97} is used as the sentence encoder, and the size of hidden state is $200$. A 5-layer Transformer encoder \cite{DBLP:conf/nips/VaswaniSPUJGKP17} with $4$ heads is used as the document-level self-attention module. A linear classification layer is used to predict whether to choose the sentence.

The training process consists of two phrases. First, we use the pre-training task to pre-train the basic model using the raw article from the CNN/DM dataset without labels. Second, we fine-tune the pre-trained model for the extractive summarization task using the sentence labels. The learning rate is set as $0.0001$ in the pre-training phase and $0.00001$ in the fine-tune phase. We train each pre-training task until it is converged or the number of training epochs reaches the upper bound $30$. We set the probability to mask, replace or switch sentences as $0.25$.

\begin{table}
    \centering
    \begin{tabular}{cccc}
    \toprule
    \textbf{Method}    & \textbf{Rouge-1} & \textbf{Rouge-2} & \textbf{Rouge-L} \\\midrule
    Basic & 41.07 & 18.95 & 37.56\\\midrule
    LEAD3 & 39.93 & 17.62 & 36.21 \\
    \textsc{NeuSum} & {41.18}$^*$ & 18.84 & {37.61} \\
    \midrule
    Mask & 41.15$^*$ & 19.06$^*$ & 37.65$^*$ \\
    Replace & 41.21$^*$ & 19.08$^*$ & 37.73$^*$ \\
    Switch & \bf{41.36}  & \bf{19.20} & \bf{37.86}  \\\midrule
    SentEnc & 41.17$^*$ & 19.04$^*$ & 37.69$^*$ \\
    Switch 0.15 & 41.35$^*$ & 19.18$^*$ & 37.85$^*$ \\
    Switch 0.35 & 41.27$^*$ & 19.12$^*$ & 37.77$^*$\\
    \bottomrule
    \end{tabular}
    \caption{
    The Rouge \cite{lin2004rouge} scores for the basic model, baselines, pre-training methods, and analytic experiments. All of our Rouge scores have a $95\%$ confidence interval of at most $\pm0.25$ as reported by the official ROUGE script. The best result is marked in bold, and those that are not significantly worse than the best are marked with $^*$.}
    \label{tab:rouge_score}
\end{table}

\paragraph{Results} We show the Rouge score on the development set during the training process in Figure \ref{rouge score fig}, and present the best Rouge score for each method in Table \ref{tab:rouge_score}.
All pre-training methods improve the performance compared with the Basic model.
Especially, Switch method achieves the best result on all the three evaluations compared with other pre-training methods, and is even better than the state-of-the-art extractive model \textsc{NeuSum}\footnote{We use code from \url{https://github.com/magic282/NeuSum} to train the model, and evaluate it using our evaluation script. Results using their script (only include Rouge-1 and Rouge-2) is put in Appendix \ref{tab:rouge_score_neusum}.}. 

In the terms of convergence, the Mask, Replace and Switch task takes $21, 24, 17$ epochs in the training phase respectively, and $18, 13, 9$ epochs to achieve the best performance in the fine-tune phase. The basic model takes $24$ epochs to obtain the best result. 
From Figure \ref{rouge score fig}, we can see that the \textit{Switch} task converges much faster than the basic model. Even adding on the epochs taken in the pre-training phase, Switch method ($26$ epochs) takes roughly the same time as the Basic model ($24$ epochs) to achieve the best performance.

\subsection{Ablation Study}

\paragraph{Reuse only the sentence encoder} Our basic model has mainly two components: a sentence encoder and a document-level self-attention module. The sentence encoder focuses on each sentence, while document-level self-attention module incorporates more document information. To investigate the role of the document-level self-attention module, we only reuse the sentence encoder of the pre-train model, and randomly initialize the document-level self-attention module. The results is shown in Table \ref{tab:rouge_score} as SentEnc. We can see that using the whole pre-training model (Switch $0.25$) can achieve better performance, which indicates the model learn some useful document-level information from the pre-training task. We notice that only using the sentence encoder also get some improvement over the basic model, which means that the pre-training task may also help to learn the independent sentence representation.

\paragraph{On the sensitivity of hyper-parameter} In this part, we investigate the sensitivity of the model to the important hyper-parameter $P_w$, i.e., the probability to switch sentences. In the previous experiment, we switch sentences with probability $0.25$. We further try the probability of $0.15$ and $0.35$, and show the results in Table \ref{tab:rouge_score} as Switch $0.15$ and Switch $0.35$. We can see Switch $0.15$ achieve basically the same result as Switch $0.25$, and Switch $0.35$ is slightly worse. So the model is not so sensitive to the hyper-parameter of the probability to switch sentences, and probability between $0.15$ and $0.25$ should be able to work well.


\section{Conclusion}

In this paper, we propose three self-supervised tasks to force the model to learn about the document context, which will benefit the summarization task. Experiments on the CNN/DM verify that through the way of pre-training on our proposed tasks, the model can perform better and converge faster when learning on the summarization task. Especially, through the Switch pre-training task, the model even outperforms the state-of-the-art method \textsc{NeuSum} \cite{DBLP:conf/acl/ZhaoZWYHZ18}. Further analytic experiments show that 
the document context learned by the document-level self-attention module will benefit the model in summarization task, and the model is not so sensitive to the hyper-parameter of the probability to switch sentences.

\bibliography{acl2019}
\bibliographystyle{acl_natbib}

\appendix

\clearpage

\section{Appendix}

\begin{figure}[!h]
\centering
\begin{minipage}{0.99\linewidth}\centering
\includegraphics[width=7cm]{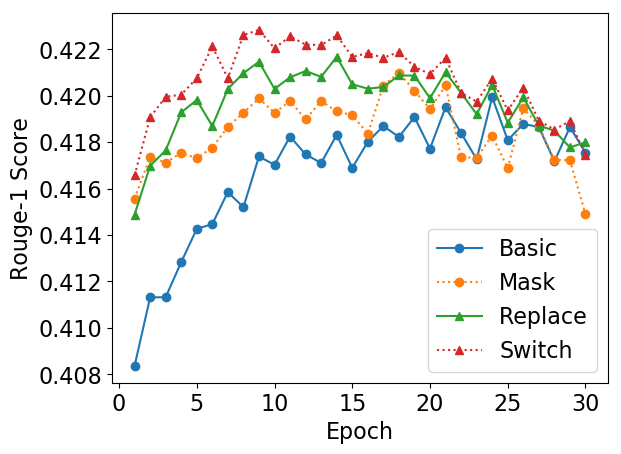}\\
(a) Rouge-1
\end{minipage}\\
\begin{minipage}{0.99\linewidth}\centering
\includegraphics[width=7cm]{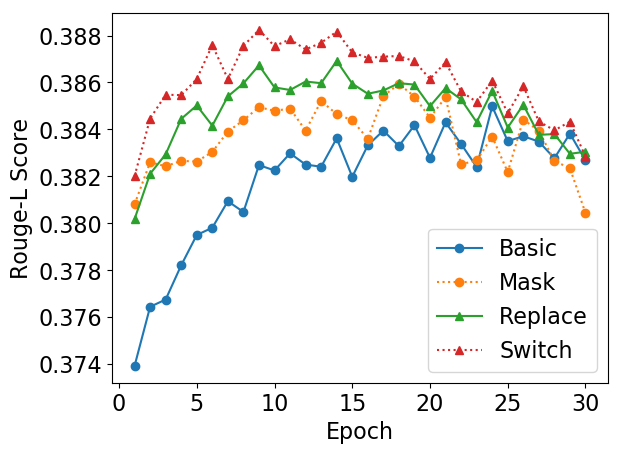}\\
(b) Rouge-L
\end{minipage}
\caption{The Rouge-1 and Rouge-L score for each pre-training method and the basic model on the development set during the training process.}
\label{rouge-1 and rouge-L}
\end{figure}

\subsection{Evaluation results using scripts from \textsc{NeuSum}}
\label{tab:rouge_score_neusum}
\begin{table}[!h]
    \centering
    \begin{tabular}{ccc}
    \toprule
    Method    & Rouge-1 & Rouge-2  \\\midrule
    Basic & 41.13 & 18.97\\\midrule
    Mask & 41.21$^*$ & 19.07$^*$  \\
    Replace & 41.27$^*$ & 19.09$^*$\\
    Switch & \bf{41.41}  & \bf{19.22}  \\\midrule
    LEAD3 & 39.98 & 17.63 \\
    \textsc{NeuSum}$^-$ & {41.23}$^*$ & 18.85  \\
    \bottomrule
    \end{tabular}
    \caption{The Rouge \cite{lin2004rouge} score for basic model, the pre-training methods, and the baselines. We use the script from \url{https://github.com/magic282/NeuSum} to compute the Rouge score. All of our Rouge scores have a $95\%$ confidence interval of at most $\pm0.22$ as reported by the official ROUGE script. The best result for each score is marked in bold, and those that are not significantly worse than the best are marked with $^*$.}
\end{table}

\subsection{Rouge-1 and Rouge-L results}
\label{rouge-1-L}
The Rouge-1 and Rouge-L results are shown in Figure \ref{rouge-1 and rouge-L}, from which we can see that the Switch method achieves the best performance.

\end{document}